\documentclass[twoside]{article}

\usepackage{my_modified}
\usepackage{algorithm}
\usepackage{algpseudocode}
\usepackage{amsmath}
\usepackage{multirow}
\usepackage{graphicx}
\usepackage{amsthm}
\usepackage[round]{natbib}

\usepackage{amssymb}
\usepackage{mathtools}

\newtheorem{theorem}{Theorem}
\newtheorem{theorem1}{Theorem}
\newtheorem{proposition}{Proposition}
\newtheorem{property}{Property}

\begin{document}



\twocolumn[
\aistatstitle{Generalizing and Improving Jacobian and Hessian Regularization}
\aistatsauthor{ Chenwei Cui \And Zehao Yan \And  Guangshen Liu \And Liangfu Lu \mbox{*} }
\aistatsaddress{ Boston University \And  Ohio State University \And Tianjin University \And Tianjin University}
]

\begin{abstract}
Jacobian and Hessian regularization aim to reduce the magnitude of the first and second-order partial derivatives with respect to neural network inputs, and they are predominantly used to ensure the adversarial robustness of image classifiers. In this work, we generalize previous efforts by extending the target matrix from zero to any matrix that admits efficient matrix-vector products. The proposed paradigm allows us to construct novel regularization terms that enforce symmetry or diagonality on square Jacobian and Hessian matrices. On the other hand, the major challenge for Jacobian and Hessian regularization has been high computational complexity. We introduce Lanczos-based spectral norm minimization to tackle this difficulty. This technique uses a parallelized implementation of the Lanczos algorithm and is capable of effective and stable regularization of large Jacobian and Hessian matrices. Theoretical justifications and empirical evidence are provided for the proposed paradigm and technique. We carry out exploratory experiments to validate the effectiveness of our novel regularization terms. We also conduct comparative experiments to evaluate Lanczos-based spectral norm minimization against prior methods. Results show that the proposed methodologies are advantageous for a wide range of tasks.
\end{abstract}

\section{Introduction}
Regularizing the Jacobian and Hessian matrices of neural networks with respect to inputs have long been of interest due to their connection with the generalizability and adversarial robustness of neural networks \citep{doubleBackprop, jacRegFst, hessRegGD}.
However, exact construction of the Jacobian and Hessian matrices is expensive. 
The computational cost scales linearly with the dimensionality of network inputs and outputs \citep{cheapDiffOP, hessRegGD}.
Early attempts to alleviate this difficulty either are not able to scale up to neural networks with high dimensional inputs and outputs, or rely on cumbersome designs while still having large estimation variances \citep{doubleBackprop, jacRegLayerwiseApprox, curvProp}.

With the emergence of vector-Jacobian product (VJP) \citep{pytorch2017}, Jacobian-vector product (JVP) \citep{JVPOri}, and Hessian-vector product (HVP) \citep{fastHVP}, recent efforts have turned to well-established matrix-free methods.
Such methodologies are more principled and elegant in that they reuse existing theories and are backed with mature implementations.

One class of these works uses Hutchinson’s trace estimator \citep{hutchinson} to construct unbiased estimates for quantities such as the Frobenius norm of Jacobians or Hessians \citep{jacRegFst, jacRegSnd, slicedSM}. However, such methods endure large variances that hinder training and generalization due to the stochastic nature of Hutchinson's estimator.

A recently emerging line of research instead focuses on minimizing the spectral norms of Jacobians and Hessians \citep{jacRegPM, hessRegGD}.
The rationale is two-fold. For one thing, due to the equivalence of norms, reducing spectral norms minimizes Frobenius norms. For another, spectral norms can be accurately obtained for a constant computational cost \citep{eig20th}. However, current efforts still rely on rudimentary algorithms such as Power Method or gradient ascent to calculate the spectral norms, overlooking the mature line of research for eigenvalue problems. In fact, existing research strongly indicates that Lanczos algorithm is ideal for this task \citep{lanczosCC, eig20th}.

Another unsatisfactory phenomenon we observe is the lack of flexibility: most existing works focus on training Jacobians and Hessians into zero \citep{doubleBackprop, jacRegFst, hessRegGD}, and few have explored the possibility of training them into arbitrary matrices, much less matrices with certain properties, such as symmetry and diagonality. As we later discuss in Sec.~4.1, enforcing symmetry or diagonality upon square Jacobian and Hessian matrices has important implications for Energy-based Models (EBMs) \citep{shouldEBM} and generative models \citep{hessPenalty}.

In this work, we first generalize the regularization of Jacobians and Hessians, allowing for the conformation to arbitrary target matrices or matrices with certain properties. Next, we propose Lanczos-based spectral norm minimization, an improved methodology to optimize the regularization terms.

We start by deriving conditions under which a target matrix can be conformed to. Following the conditions, we propose novel regularization terms that match a Jacobian or Hessian matrix with a function of itself. We show that the proposed regularizer can enforce symmetry and diagonality upon square Jacobian and Hessian matrices of neural networks.

To reliably optimize the proposed regularization terms, we implement a parallelized version of the Lanczos algorithm \citep{lanczosCC}. We provide the detail of the algorithm and explain how to perform the subsequent spectral norm minimization.

To validate the effectiveness of our proposed regularizers, we construct exploratory high-dimensional tasks that are detailed in Sec.~4.1. We observe strong results that adhere to our theoretical analyses.

To rigorously compare our Lanczos-based spectral norm minimization with previous methodologies, we present extensive controlled experiments in the context of adversarial robustness. We implement all methodologies ourselves to ensure a rigorous and fair comparison.
We use ResNet-18 \citep{resnet} and CIFAR-10 and CIFAR-100 datasets \citep{cifar10100}.
Strong and standard adversary, namely PGD(20), is used to evaluate the performance. A running time analysis is also conducted for our technique. The experiments show that our Lanczos-based spectral norm minimization not only is efficient to compute but also surpasses prior methods by a large margin, in terms of performance.

To summarize our main contributions:
\begin{itemize}
\item We generalize the task of regularizing Jacobians and Hessians of neural networks with respect to inputs, permitting arbitrary target matrices.
\item We explore novel training objectives that enforce symmetry or diagonality for square matrices, which are validated theoretically and empirically. It opens up new possibilities for applications.
\item We propose Lanczos-based spectral norm minimization, an effective technique for Jacobian and Hessian training. Not only being theoretically sound, experiments also show evident improvements over prior methods.
\end{itemize}

\textbf{Notation.} We summarize the notations used throughout this paper. By convention, we use regular letters for scalars and bold letters for both vectors and matrices. Neural networks are denoted by function $f(\mathbf{x};\mathbf{\theta})$, which can have single or multiple outputs with respect to the specific situation. The Jacobian matrix of $f$ at point $\mathbf{x}$ is denoted by $\mathbf{J}_f(\mathbf{\theta};\mathbf{x})$, and the Hessian matrix of $f$ at point $\mathbf{x}$ is denoted by $\mathbf{H}_f(\mathbf{\theta};\mathbf{x})$. When we talk about Jacobian or Hessian matrices, we use $\mathbf{A}(\mathbf{\theta};\mathbf{x})$ to denote $\mathbf{J}$ or $\mathbf{H}$. In some circumstances, for simplicity, we omit the inputs. Instead, we denote them as $\mathbf{J}_f,\mathbf{H}_f,\mathbf{A}_f$. For any vector $\mathbf{x}$, we denote $\Vert \mathbf{x}\Vert_2$ as its L2 norm. For any matrix $\mathbf{A}$, $\mathbf{A}^T$ means its transpose, its Frobenius norm is denoted by $\Vert \mathbf{A}\Vert_F$, and its spectral norm is denoted by $\Vert \mathbf{A}\Vert_2$. $\mathrm{Tr}(\mathbf{A})$ denotes the trace of $\mathbf{A}$. $\mathbf{I}$ denotes the unit matrix. $\mathbf{A}_{ij}$ means the $i,j$ entries of matrix $\mathbf{A}$. $\sigma_{\max}(\mathbf{A})$ denotes its largest singular value. $\lambda_{\max}(\mathbf{A})$ denotes its largest eigenvalue, and its corresponding unit eigenvector is denoted by $\mathbf{v}_m$.

\section{Related Work}
\textbf{Early efforts} focused on training the Jacobians of neural networks with respect to inputs trace back to \citet{doubleBackprop}. The authors propose double propagation to regularize the Jacobians of loss functions. However, this algorithm only applies to computational graphs with single outputs. For a more general case, \citet{jacRegLayerwiseApprox} utilize layer wise approximations to regularize the Jacobians of neural networks with multiple inputs and outputs. For Hessians, \citet{regSM} reduce the cost of backpropagation by limiting it to differentiate the diagonal of a Hessian matrix. \cite{curvProp} later introduce curvature propagation, an algorithm to produce stochastic estimates for Hessian matrices.

Those earlier attempts either are not able to scale up to neural networks with high dimensional inputs and outputs or rely on cumbersome design while still having large estimation variances.

Recent attempts have turned to well-established matrix-free techniques and are either based on Hutchinson’s estimators or spectral norm minimization.

\textbf{Hutchinson’s estimator} \citep{hutchinson} takes the form $\mathrm{E}[\mathbf{v}^T \mathbf{A} \mathbf{v}] = \mathrm{Tr}(\mathbf{A})$, where $\mathbf{A}$ is an arbitrary square matrix and $\mathbf{v}$ is a random vector such that $\mathrm{E}[\mathbf{v}\mathbf{v}^T] = \mathbf{I}$.
It follows when ${\mathbf{v} \sim \mathcal{N}(\mathbf{0}, \mathbf{I})}$,
\begin{equation}
\mathrm{Var}[\mathbf{v}^T\mathbf{A} \mathbf{v}] = 2\lVert \mathbf{A} \rVert _F^2.
\end{equation}
Instead, when $\mathbf{v}$ is drawn from a multivariate Rademacher distribution,
\begin{equation}
\mathrm{Var}[\mathbf{v}^T\mathbf{A}\mathbf{v}] = 2\sum_i\sum_{j\neq i}\mathbf{A}_{ij}^2.
\end{equation}

\citet{jacRegFst} first propose to use random projections to regularize the Jacobian $\mathbf{J}_f(\mathbf{x}; \theta)$ of a neural network.
The same technique is later revisited by \cite{jacRegSnd}.
Specifically, given $\mathbf{v} \sim \mathcal{N}(\mathbf{0}, \mathbf{I})$, $\lVert \mathbf{v}^T \mathbf{J}_f \rVert_2^2$ is minimized.
This is an instance of Hutchinson's estimators in that
$$
\mathrm{E}[\lVert \mathbf{v}^T \mathbf{J}_f \rVert_2^2] = \mathrm{E}[\mathbf{v}^T \mathbf{J}_f \mathbf{J}_f^T \mathbf{v}] = \mathrm{Tr}(\mathbf{J}_f \mathbf{J}_f^T) = \lVert \mathbf{J}_f \rVert_F^2.
$$
Consequently, the variance is a significant $2\lVert \mathbf{J}_f \mathbf{J}_f^T \rVert_F^2$.

For Hessians, \citet{slicedSM} propose sliced score matching, in which Hutchinson’s estimators are used to maximize the trace of Hessians. However, sliced score matching is often observed to be too stochastic and less performant, compared with its Hessian-free counterparts \citep{denoisingSM}.

Another significant adoption of Hutchinson's estimator is the Hessian penalty \cite{hessPenalty}.
The authors propose unbiased estimators to regularize off-diagonal elements of Hessians.
Essentially, this technique is built upon Eq. (2).
Nonetheless, this estimator endures high variance since, in practice, the authors use empirical variance, calculated from only two samples.

Being unbiased estimators, Hutchinson-based methods are theoretically sound. However, in practice, the variance of these estimators is significant and reduces performance. (we validate in experiments)

\textbf{Spectral norm minimization} is a recently emerging line of research that instead focuses on minimizing the spectral norms of Jacobians and Hessians.
Spectral norms can be accurately obtained at a constant cost \citep{eig20th}, and has the norm equivalence
$$\lVert \mathbf{A} \rVert_2 \leq \lVert \mathbf{A} \rVert_F \leq \sqrt{r} \lVert \mathbf{A} \rVert_2,$$
for any matrix $\mathbf{A}$ of rank $r$.

Input Hessian regularization \citep{hessRegGD} considers the term $\lVert \mathbf{H}_f\mathbf{v} \rVert_2$ and uses gradient ascent to solve for $v$, in an attempt to find the spectral norm of $\mathbf{H}_f$.
We however show in Appendix A that this method is closely related to power iteration.
In certain cases, they are outright equivalent.
However, power iteration generally converges much slower than Lanczos algorithm. Depending on the matrix, it may even cease to converge \citep{eig20th}.
Since both methods have computational costs dominated by matrix-vector products and therefore take up similar running times, it is hard to justify using power iteration instead of Lanczos algorithm.

For Jacobians, a concurrent work \citep{jacRegPM} recently proposes to use power iteration to find spectral norms.
However, as aforementioned, convergence of power iteration is slow and not guaranteed.

Research regarding spectral norm minimization is still at an early stage.
Lanczos-based spectral norm minimization not only is theoretically sound but also empirically surpasses existing methods by a large margin (see Sec.~4.4).

\section{Methodology}
\subsection{Spectral Norm Minimization}
We start our exposition by formulating the problem of training Jacobian and Hessian matrices into zero, using spectral norm minimization.
Subsequently, we outline conditions under which spectral norm minimization can be performed.

We consider a matrix $\mathbf{A}_f(\mathbf{x}; \mathbf{\theta})$.
It can either be a Jacobian or a Hessian matrix resulting from a neural network.
Our exposition does not depend on the particular width and height of $\mathbf{A}_f(\mathbf{x}; \mathbf{\theta})$.
Specifically, we make a trivial assumption that the neural networks are smooth and in turn their Hessians are symmetric. In our experiments, we use Softplus activation function \citep{softplus} to ensure smoothness.

Given $\mathbf{A}_f(\mathbf{x}; \mathbf{\theta})$, to train it into a zero matrix, the common idea is to minimize its Frobenius norm,
$$\lVert \mathbf{A}_f(\mathbf{x}; \mathbf{\theta}) \rVert_F = \sqrt{\sum_i \sum_j [\mathbf{A}_f(\mathbf{x}; \mathbf{\theta})]_{i j} ^2}.$$
However, direct minimization of $\lVert \mathbf{A}_f(\mathbf{x}; \mathbf{\theta}) \rVert_F$ requires an exact construction.
This is usually impractical for the Jacobians and Hessians of neural networks with high-dimensional inputs or outputs.

Fortunately, by minimizing the spectral norm $\lVert \mathbf{A}_f \rVert_2$, $\lVert \mathbf{A}_f \rVert_F$ can be properly minimized. Consider the following norm equivalence:
$$\lVert \mathbf{A}_f\rVert_2 \leq \lVert \mathbf{A}_f \rVert_F \leq \sqrt{r} \lVert \mathbf{A}_f \rVert_2,$$ where $r$ is the rank of matrix $\mathbf{A}_f.$
It shows that as $\lVert \mathbf{A}_f \rVert_2 \rightarrow 0$, $\lVert \mathbf{A}_f\rVert_2$ becomes an increasingly tight bound.
Also, $\lVert \mathbf{A}_f \rVert_F = 0$ if and only if $\lVert \mathbf{A}_f \rVert_2 = 0$.
Therefore, we minimize $\lVert \mathbf{A}_f\rVert_2$ instead.

We take notice that the spectral norm of $\mathbf{A}_f$ is the maximum singular value $\sigma_{\max} (\mathbf{A}_f)$, which is by definition equal to $\sqrt{\lambda_{\max} (\mathbf{A}_f \mathbf{A}_f^T)}$, where $\lambda_{\max}(\mathbf{A}_f \mathbf{A}_f^T)$ is the maximum eigenvalue of $\mathbf{A}_f \mathbf{A}_f^T$ in terms of magnitude.
It is convenient to note that $\mathbf{A} \mathbf{A}^T$ is symmetric and positive semi-definite, for any matrix $\mathbf{A}$.
Also, since $\mathbf{A} \mathbf{A}^T$ is symmetric, we have
$$\lambda_{\max} (\mathbf{A} \mathbf{A}^T) = \mathbf{v}_m^T \mathbf{A} \mathbf{A}^T \mathbf{v}_m = \lVert \mathbf{v}_m^T \mathbf{A}  \rVert_2,$$
where $\mathbf{v}_m$ is the normalized eigenvector corresponding to $\lambda_{\max} (\mathbf{A} \mathbf{A}^T)$.

So far we have transformed spectral norm minimization into minimizing  $\sqrt{\lambda_{\max} (\mathbf{A}_f \mathbf{A}_f^T)}$.
For this purpose, we take two steps: 1) We obtain $\mathbf{v}_m$ by solving the extremal eigenvalue problem for $\mathbf{A}_f \mathbf{A}_f^T$. 2) Given $\mathbf{v}_m$, we minimize $\lVert \mathbf{v}_m^T \mathbf{A}_f  \rVert_2$.

For 1), we use Lanczos algorithm to solve for $\mathbf{v}_m$.
For now, we focus on the conditions that $\mathbf{A}_f$ should satisfy, and elaborate other details in (Sec.~3.4).
Lanczos algorithm operates on Hermitian matrices and requires the matrices to admit efficient matrix-vector products. The first condition is met since $\mathbf{A} \mathbf{A}^T$ is always symmetric. The second condition requires the existence of an efficient $\mathbf{A} \mathbf{A}_f^T \mathbf{v}$ operator.

For 2), we minimize $\lVert \mathbf{v}_m^T \mathbf{A}  \rVert_2$, given $\mathbf{v}_m$. For Jacobians and Hessians, the minimization is made possible by VJP, JVP, and HVP. It follows that $\mathbf{A}_f$ should permit an efficient vector-matrix product operator.

The following proposition summarizes the conditions under which $\mathbf{A}_f$ can be optimized by spectral norm minimization.
\begin{proposition}
Matrix $\mathbf{A}_f$ can be optimized by spectral norm minimization if both of the following satisfies: \\
1) $\forall{\mathbf{v}}, \mathbf{A}_f \mathbf{A}_f^T \mathbf{v}$ can be efficiently computed. \\
2) $\forall{\mathbf{v}}, \mathbf{v}^T \mathbf{A}_f$ can be efficiently computed.
\end{proposition}

We conclude this section by quickly validating the satisfiability of the conditions for Jacobian and Hessian matrices of neural networks.

For a Jacobian $\mathbf{J}_f$, we have $\mathbf{J}_f \mathbf{J}_f^T \mathbf{v} = \mathbf{J} (\mathbf{v}^T \mathbf{J})^T, \forall{\mathbf{v}}$. It can be efficiently computed using VJP and JVP operators. Also, by the definition of VJP, we can efficiently compute $\mathbf{v}^T \mathbf{J}_f$.

For a Hessian $\mathbf{H}_f$, we note that it is symmetric since we assume smooth neural networks. Therefore, we have $\forall{\mathbf{v}}, \mathbf{H}_f \mathbf{H}_f^T \mathbf{v} = \mathbf{H}_f(\mathbf{H}_f\mathbf{v})$ and $\mathbf{v}^T \mathbf{H}_f = \mathbf{H}_f\mathbf{v}$. Both can be efficiently obtained given the HVP operator.

\subsection{Generalized Jacobian and Hessian Regularization}
In this section, we generalize the idea of $\lVert \mathbf{A}_f \rVert_F$ minimization. The new paradigm allows training a matrix $\mathbf{A}$ into any arbitrary target matrix $\mathbf{A}_0$, hence the name generalized Jacobian and Hessian regularization. Further, we derive conditions that $\mathbf{A}_0$ should follow in order to be a valid target for spectral norm minimization.

To conform $\mathbf{A}_f$ to $\mathbf{A}_0$, it is straightforward to minimize $\lVert \mathbf{A}_f - \mathbf{A}_0 \rVert_F$. We can efficiently minimize $\lVert \mathbf{A}_f - \mathbf{A}_0 \rVert_F$ using spectral norm minimization, as long as $\mathbf{A}_f - \mathbf{A}_0$ follows proposition 1.

To begin with, we consider $(\mathbf{A}_f - \mathbf{A}_0) (\mathbf{A}_f - \mathbf{A}_0)^T \mathbf{v}, \forall{\mathbf{v}}$. Expanding it gives us:
\begin{align*}
	&(\mathbf{A}_f - \mathbf{A}_0) (\mathbf{A}_f - \mathbf{A}_0)^T \mathbf{v} \\
	=& \mathbf{A}_f\mathbf{A}_f^T\mathbf{v} -\mathbf{A}_f\mathbf{A}_0^T\mathbf{v} - \mathbf{A}_0\mathbf{A}_f^T\mathbf{v} + \mathbf{A}_0\mathbf{A}_0^T\mathbf{v}
\end{align*}

We can therefore conclude the following proposition.

\begin{proposition}
Matrix $\mathbf{A}_0$ is an valid target matrix for spectral norm minimization if both of the following satisfies: \\
1) $\forall{\mathbf{v}}, \mathbf{v}^T \mathbf{A}_0$ can be efficiently computed. \\
2) $\forall{\mathbf{v}}, \mathbf{A}_0 \mathbf{v}$ can be efficiently computed.
\end{proposition}

Proposition 2 implies that any matrix that permits an efficient left and right vector product is a valid target matrix. This ensures flexibility when choosing $\mathbf{A}_0$. For example, $\mathbf{A}_0$ can be an explicit constant matrix, the Jacobian or Hessian resulting from another neural network, or any transformation of $\mathbf{A}_f$ that preserves vector products (see Sec.~3.3).

\subsection{Enforcing Symmetric or Diagonal Matrices}
In this section, we propose the novel observation that we can enforce certain properties upon Jacobian and Hessian matrices, using spectral norm minimization. Specifically, we propose formulas that enforce symmetry or diagonality for Jacobian and Hessian matrices of neural networks, with respect to network inputs.

\textbf{Symmetry.}
For symmetry, we consider Jacobians $\mathbf{J}_f$ of neural networks whose number of inputs equals the number of outputs. In this case, Jacobians are square matrices but are generally non-symmetric \citep{shouldEBM}.

By the definition of symmetry, we expect $\mathbf{J}_f = \mathbf{J}_f^T$. An accurate depiction of this objective is $\lVert \mathbf{J}_f - \mathbf{J}_f^T \rVert_F$ minimization. We soon notice that by making $\mathbf{J}_f^T$ the target matrix, it is possible to enforce symmetry for square Jacobians. It is easy to validate that $\mathbf{J}_f^T$ satisfies proposition 2, given VJP and JVP operators. Therefore, we can indeed optimize $\lVert \mathbf{J}_f - \mathbf{J}_f^T \rVert_F$ efficiently using spectral norm minimization.

In practice, to find the spectral norm, we provide
$$(\mathbf{J}_f - \mathbf{J}_f^T) (\mathbf{J}_f - \mathbf{J}_f^T\mathbf{)}^T\mathbf{v} = \mathbf{J}_f\mathbf{J}_f^T\mathbf{v} -\mathbf{J}_f\mathbf{J}_f\mathbf{v} - \mathbf{J}_f^T\mathbf{J}_f^T\mathbf{v} + \mathbf{J}_f^T\mathbf{J}\mathbf{v}$$
to our parallelized Lanczos algorithm.
For optimization, we simply calculate
$$\lVert \mathbf{v}_m^T (\mathbf{J}_f - \mathbf{J}_f^T)  \rVert_2 = \lVert \mathbf{v}_m^T \mathbf{J}_f - \mathbf{v}_m^T \mathbf{J}_f^T  \rVert_2$$
given eigenvector $\mathbf{v}_m$.

Sec.~4.3 presents empirical evidence that this technique is possible and efficient. The potential application for this objective includes ensuring conservative vector fields for Energy-based Models (EBMs). We elaborate more in Sec.~4.1.

\textbf{Diagonality.} For diagonality, we consider a matrix $\mathbf{A}_f$ that can either be the Jacobian or the Hessian of a neural network. The only restriction we make is that $\mathbf{A}_f$ should be a square matrix.

By the definition of diagonality, we should train all off-diagonal elements of $\mathbf{A}_f$ into zero. This objective can be described as training $ \sum_i \sum_{j \neq i} [\mathbf{A}_f(\mathbf{x};\mathbf{\theta})]_{ij}^2 $ to zero. On first sight, spectral norm minimization is not applicable to this problem. However, we propose the following theorem.

\begin{theorem}
$\forall{\mathbf{A}} \in \mathbb{R}^{N \times N}$, the following holds
$$\frac{1}{N} \lVert \mathbf{A}-\mathcal{D}(\mathbf{A}\mathbf{1})\rVert_F^2 \leq \sum_i \sum_{j \neq i} \mathbf{A}_{ij}^2 \leq \lVert \mathbf{A}-\mathcal{D}(\mathbf{A}\mathbf{1})\rVert_F^2,$$
where $\mathbf{1} \in \mathbb{R}^{N}$ is an all-one vector, and $\mathcal{D}$ is a function that transforms a vector into a diagonal matrix.
\end{theorem}

The proof of the above theorem is provided in Appendix B.

Theorem 1 shows that as $\lVert \mathbf{A}_f-\mathcal{D}(\mathbf{A}_f\mathbf{1})\rVert_F^2 \rightarrow 0$, $\lVert \mathbf{A}_f-\mathcal{D}(\mathbf{A}_f\mathbf{1})\rVert_F^2$ becomes an increasingly tight bound.
Also, $\sum_i \sum_{j \neq i} [\mathbf{A}_f(\mathbf{x};\mathbf{\theta})]_{ij}^2=0$ if and only if ${\lVert \mathbf{A}_f-\mathcal{D}(\mathbf{A}_f\mathbf{1})\rVert_F^2 = 0}$.
We therefore minimize ${\lVert \mathbf{A}_f-\mathcal{D}(\mathbf{A}_f\mathbf{1})\rVert_F^2}$ instead.

Next, we validate the satisfiability of proposition 2 for $\mathcal{D}(\mathbf{A}_f\mathbf{1})$. We first consider the following property of $\mathcal{D}$.
\begin{property}
$\forall{\mathbf{v}_1, \mathbf{v}_2} \in \mathbb{R} ^{N}, \mathbf{v}_2^T \mathcal{D}(\mathbf{v}_1) = \mathcal{D}(\mathbf{v}_1) \mathbf{v}_2 = \mathbf{v}_1 \odot \mathbf{v}_2$
\end{property}
It follow that
$$\forall{\mathbf{v}}, \mathbf{v}^T \mathcal{D}(\mathbf{A}_f\mathbf{1}) = \mathcal{D}(\mathbf{A}_f\mathbf{1}) \mathbf{v} = (\mathbf{A}_f\mathbf{1}) \odot \mathbf{v}.$$
Therefore, we can optimize $\lVert \mathbf{A}_f-\mathcal{D}(\mathbf{A}_f\mathbf{1})\rVert_F$ efficiently using spectral norm minimization.

Sec.~4.3 presents empirical evidence that we can efficiently enforce diagonality for Hessians. The potential application for this objective includes performing disentanglement for deep generative models. We elaborate the background in Sec.~4.1.

\subsection{Lanczos-Based Spectral Norm Minimization}
In this section, we focus on the extremal eigenvalue problem. Specifically, given a symmetric matrix $\mathbf{A}$, we want to find the largest eigenvalue $\lambda_{\max}(\mathbf{A})$ in terms of magnitude and its corresponding normalized eigenvector $v_m$. For this purpose, we introduce our implementation of the parallelized Lanczos algorithm.

Given a batch of square matrices, denoted by $\mathbf{A} \in \mathbb{R} ^ {b \times d \times d}$, where $b$ is the batch size, and $d$ is the dimensionality of the square matrices. We construct the batched matrix-vector product function $\mathcal{M}: \mathbb{R} ^ {b \times d} \rightarrow \mathbb{R} ^ {b \times d}$. For a batch of vectors, $\mathcal{M}$ computes the matrix-vector products in a parallel manner.

We propose Algorithm 1, the parallelized Lanczos algorithm.
After computation, Algorithm 1 returns a batch of tridiagonal matrices $\mathbf{T}$ and a tensor consisting of Lanczos vectors $\mathbf{V}$.
To obtain the normalized eigenvectors corresponding to the largest eigenvalues, we first compute the eigenvalues and eigenvectors of $\mathbf{T}$ using traditional batched eigensolvers \citep{pytorch2017}. Since the width of each tridiagonal matrix is exactly the iteration number $n$, the computation is negligible. Moreover, the eigenvalues of $\mathbf{T}$ are the same as the real eigenvalues. Afterwards, $\mathbf{V}$ can be used to map the eigenvectors resulting from $\mathbf{T}$ to the actual eigenvectors.
Through this procedure, accurate extremal eigenvalues and eigenvectors can be obtained, at the cost of only a few iterations.

A running time analysis of this algorithm is performed in Sec.~4.5.

\begin{algorithm}[tb]
\caption{Parallelized Lanczos Algorithm}
\label{alg:algorithm}
\textbf{Input}: \\
$\mathcal{M}$, batched matrix-vector product function.\\
$b$, batch size.\\
$n$, iteration number.\\
$d$, dimensionality.\\
\textbf{Output}:\\
$\{\mathbf{V}^{(i)}\}, i=1, ..., b$ where $\mathbf{V}^{(i)} \in \mathbb{R}^{n \times d}$. \\
$\{\mathbf{T}^{(i)}\}, i=1, ..., b$ where $\mathbf{T}^{(i)} \in \mathbb{R}^{n \times n}$. \\
\begin{algorithmic}[1] %[1] enables line numbers
\State Initialize $\mathbf{V}^{(i)} \in \mathbb{R}^{b \times d}, i=1, ..., n$ as zero matrices.
\State Initialize $\mathbf{T}^{(i)} \in \mathbb{R}^{b \times n}, i=1, ..., n$ as zero matrices.
\State Initialize $\mathbf{a}^{(i)} \in \mathbb{R}^b, i=1, ..., n$ as zero vectors.
\State Initialize $\mathbf{b}^{(i)} \in \mathbb{R}^b, i=1, ..., n$ as zero vectors.
\State Set the rows of $\mathbf{V}^{(0)}$ as random unit vectors.
\State $\omega \leftarrow \mathcal{M}(\mathbf{V}^{(0)})$~~~~~// batched matrix-vector product
\State $\mathbf{a}^{(0)} \leftarrow$ $\mathrm{dot}(\omega, \mathbf{V}^{(0)})$~~~~~// batched dot product
\State $\omega \leftarrow \omega - \mathbf{a}^{(0)} \mathbf{V}^{(0)}$
\For{$i = 1, 2, 3, ..., n - 1$}
\State $\mathbf{b}^{(i)} = \mathrm{norm}(\omega)$~~~~~// batched L2 norm
\State $\mathbf{V}^{(i)} \leftarrow \omega / \mathbf{b}^{(i)}$
\State Set NaN rows in $\mathbf{V}^{(i)}$ as random unit vectors.
\State $\omega \leftarrow \mathcal{M}(\mathbf{V}^{(i)})$
\State $\mathbf{a}^{(i)} \leftarrow$ $\mathrm{dot}(\omega, \mathbf{V}^{(i)})$~~~~~// batched dot product
\State $\omega \leftarrow \omega - \mathbf{a}^{(i)} \mathbf{V}^{(i)} - \mathbf{b}^{(i)} \mathbf{V}^{(i-1)}$
\EndFor

\For{$j = 0, 1, 2, 3, ..., n - 1$}
\State $\mathrm{col}_j(\mathbf{T}^{(j)}) = \mathbf{a}^{(j)}$
\If {$j \neq 0$}
\State $\mathrm{col}_{j+1}(\mathbf{T}^{(j)}) = \mathbf{b}^{(j)}$
\State $\mathrm{col}_j(\mathbf{T}^{(j+1)}) = \mathbf{b}^{(j)}$
\EndIf
\EndFor

\State Permute the first two axes of $\mathbf{V}$ s.t.
\Statex ~~~~$\mathbf{V}^{(i)} \in \mathbb{R}^{n \times d}, i=1, ..., b$
\State Permute the first two axes of $\mathbf{T}$ s.t.
\Statex ~~~~$\mathbf{T}^{(i)} \in \mathbb{R}^{n \times n}, i=1, ..., b$
\State \textbf{return}  $\mathbf{V}, \mathbf{T}$
\end{algorithmic}
\end{algorithm}

\section{Experiments}
\subsection{Tasks}
\textbf{Overview.}
We experiment on four tasks that validate different aspects of our generalized Jacobian and Hessian regularization and the Lanczos-based spectral norm minimization technique.

\textbf{Conservative Vector Field.}
Recently, Energy-based Models (EBMs) are demonstrating superior performance on tasks such as image generation \citep{EBM, shouldEBM, ncsn}. EBMs are traditionally scalar-valued functions that predict unnormalized probability distributions \citep{shouldEBM}. In contrast, recent efforts significantly improve performance by directly predicting the gradient vectors of the distributions \citep{ncsn}. This is however a paradoxical situation: vector-valued neural networks are not guaranteed to output a conservative vector field and therefore contradicts the assumptions that EBMs make \citep{shouldEBM}.

We approach this problem by first noting that a continuously differentiable vector field is conservative if and only if its Jacobian is symmetric. We consequently propose to minimize $\lVert \mathbf{J}_f - \mathbf{J}_f^T \rVert_F$ via our Lanczos-based spectral norm minimization to enforce symmetric Jacobians.

To validate this idea, we consider $N$-dimensional functions of the form
$f(x_1, ..., x_N) = \sum_{i=1}^{N}{g(x_i)},$
where $g$ is a differentiable unary function.
A feed forward neural network is used to learn the gradient field of $f$.
The data points are sampled from $\mathcal{N}(\mathbf{0}, \mathbf{I})$.
We report test time mean squared error and $\lVert \mathbf{J}_f - \mathbf{J}_f^T \rVert_F$ to demonstrate the effectiveness of our technique.

\textbf{Disentanglement.}
Disentanglement of high-dimensional functions have wide applications in the field of deep generative models \citep{hessPenalty}. \citet{hessPenalty} propose the notion of disentanglement that is achieved by enforcing diagonal Hessian matrices of a scalar function. For this purpose, the authors propose a stochastic estimator to penalize off-diagonal elements of Hessians.

Due to \mbox{Theorem 1}, we propose to minimize
${\lVert \mathbf{A}_f-\mathcal{D (\mathbf{A}_f\mathbf{1})\rVert_F}}$ for disentanglement.
To validate this technique, we construct \mbox{$N$-dimensional} functions of the form 
\mbox{$f(x_1, ..., x_N) = \sum_{i=1}^{N}{g(x_i)}.$}
$f$ naturally has a diagonal Hessian.
We use a feed forward neural network to learn the value of $f$.
The data points are sampled from $\mathcal{N}(\mathbf{0}, \mathbf{I})$.
We report test time mean squared error and $\sum_i \sum_{j \neq i} [\mathbf{A}_f(\mathbf{x};\mathbf{\theta})]_{ij}^2 $ to demonstrate the effectiveness of Theorem 1.\\

\textbf{Jacobian Regularization.}
To rigorously validate the effectiveness of our Lanczos-based spectral norm minimization technique, we conduct controlled experiments to compare with representative methods. Specifically, we implement and compare with normal training, Hutchinson's estimator, and Power Method.
Standard $\ell_\infty$ adversaries, namely PGD(20) is used to evaluate the performance.
We perform Jacobian Regularization on CIFAR-10 and CIFAR-100 datasets \citep{cifar10100} using \mbox{ResNet-18} \citep{resnet}.

\textbf{Hessian Regularization.}
Hessian regularization concerns matrices whose size is determined by the input number of neural networks. In our case, the associated Hessian matrix is 3072 by 3072 in size, which is magnitudes bigger compared with the matrices in Jacobian regularization. Therefore, in this task we validate the performance of Lanczos-based spectral norm minimization under situations where the relating matrices are large.
In particular, we implement and compare with normal training, Hutchinson's estimator, and Power Method. We use PGD(20) to evaluate the performance, and the experiments are conducted on both CIFAR-10 and CIFAR-100 datasets \citep{cifar10100} using ResNet-18 \citep{resnet}.

\begin{figure*}[h]
% \vspace{.3in}
\centerline{\includegraphics[width=\linewidth]{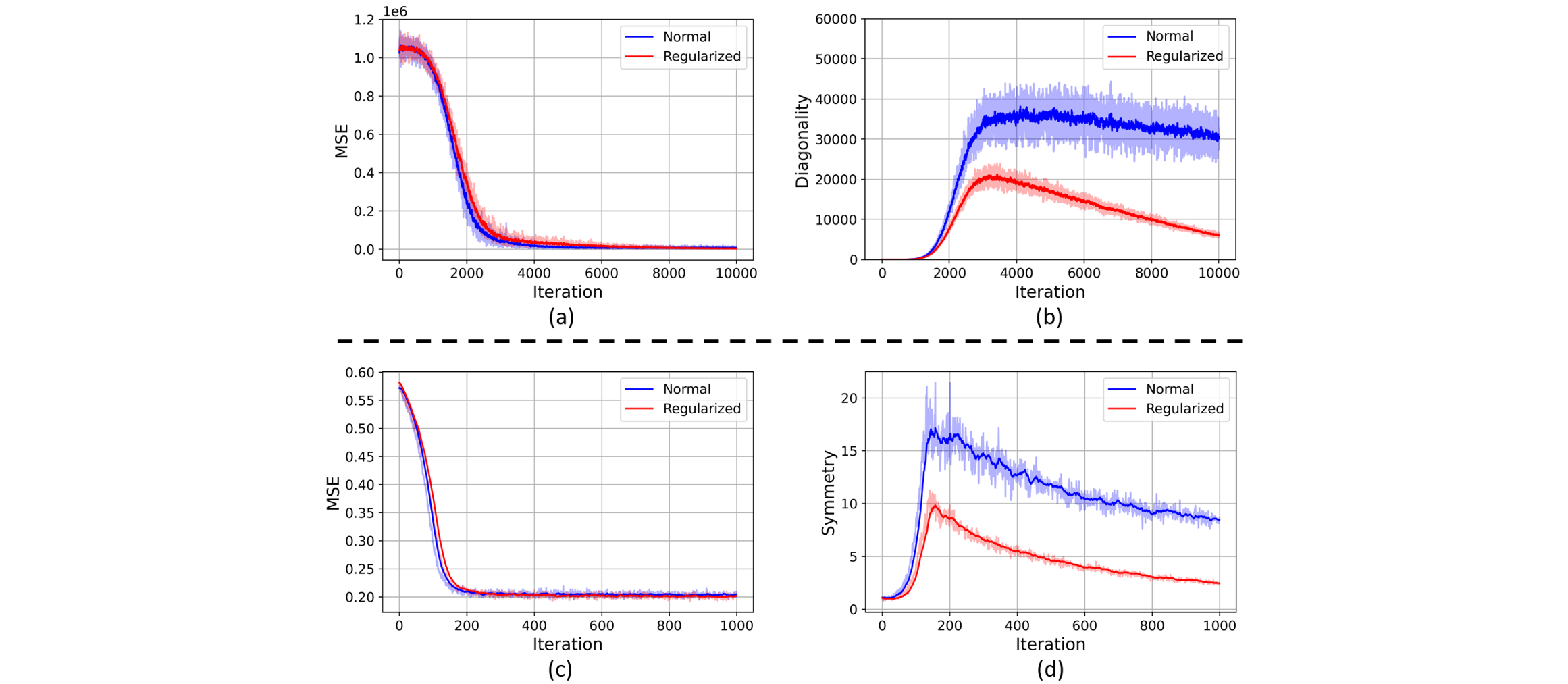}}
% \vspace{.3in}
\caption{Enforcing symmetry and diagonality using the proposed regularization terms. (a) and (b) show the results for enforcing symmetry. (c) and (d) show the results for enforcing diagonality. Symmetry is defined as ${\lVert \mathbf{J}_f - \mathbf{J}_f^T \rVert_F}$. Diagonality is defined as $\sum_i \sum_{j \neq i} [\mathbf{H}_f]_{ij}^2 .$}
\end{figure*}

\subsection{Implementation Details}

\textbf{Model Design.} We make specific design choices to ensure a simplistic implementation.
For activation functions, we use Softplus \citep{softplus} with a $\beta$ value of 8 to ensure a tight and smooth approximation to ReLU \citep{relu}.
Following \citet{VIT}, for our ResNet-18 models \citep{resnet}, we replace Batch Normalization \citep{BN} with Group Normalization \citep{GN} to avoid running statistics that may complicate our iteration-based Lanczos algorithm. Also following \citet{VIT}, standardized convolutions \citep{weightSTD} are used to accompany Group Normalization \citep{GN}.

\textbf{Hyperparameters.}
Hyperparameters are chosen according to a set of heuristics that do not favor any particular algorithm. The iteration number for Power Method and Lanczos algorithm starts with 2, and doubles each time the learning rate decays. The power of the regularizer starts from 25\% and increases 25 percentage points each time the learning rate decays, with a maximum value of 95\%. For Hutchinson's estimator only, we report an additional variant where the regularization power is further decreased by a factor of 10. This is because Hutchinson's estimator has a much greater magnitude compared with other methods and is unstable without the adjustment.

\textbf{Training.} For all experiments, the batch size is 512. We use Adam with default parameters and a starting learning rate of 0.001 for optimization. For adversarial robustness, all experiments run for 100 epochs, and the learning rate is decayed by a factor of 10 after epoch 50, 70, and 90. For CIFAR-10 and CIFAR-100 \citep{cifar10100}, random cropping and random horizontal flipping are adopted as data augmentation techniques.

\subsection{Enforcing Diagonality or Symmetry}
\textbf{Conservative Vector Field.} In this experiment, we set the function $g$ as $g(x) = \mathrm{sin}(x)$ and set the dimensionality as 1024. In Fig. 1 (a) we observe that the regularizer does not deteriorate the performance, and in Fig. 1 (b), we notice that the regularization term has a significant impact on the symmetry of the Jacobian matrix, suppressing $\lVert \mathbf{J}_f - \mathbf{J}_f^T \rVert_F$.

\textbf{Disentanglement.} In this experiment, we set the function $g$ as $g(x) = x^2$ and set the dimensionality as 1024. In Fig. 1 (c) we observe that the proposed regularization term does not decline the performance. In Fig. 1 (d), we notice that the regularizer suppresses the off-diagonal elements effectively.
This result validates the effectiveness of Theorem~1.

\begin{table} [!t]
    \begin{center}
    \caption{Experiment results for Jacobian regularization on CIFAR-10. Hutchinson-0.1 means the regularization power is reduced by a factor of 10. The results are averaged over three runs and the standard deviations are reported in parentheses.}
	\begin{tabular}{lll}
		\hline
		Method & Clean & PGD(20)\\
		\hline
		Normal & $\textbf{93.3}_{(0.1)}$ & $0.0_{(0.0)}$\\
		Hutchinson & $60.1_{(3.9)}$ & $30.1_{(1.8)}$\\
		Hutchinson-0.1 & $92.5_{(0.2)}$ & $14.2_{(0.5)}$\\
		Power Method & $75.9_{(0.1)}$ & $45.6_{(0.2)}$\\
		\textbf{Lanczos (Ours)} & $75.6_{(0.5)}$ & $\textbf{45.7}_{(0.2)}$\\
		\hline
	\end{tabular}
	\end{center}
\end{table}

\begin{table} [!t]
    \begin{center}
	\caption{Experiment results for Jacobian regularization on CIFAR-100. Hutchinson-0.1 means the regularization power is reduced by a factor of 10. The results are averaged over three runs and the standard deviations are reported in parentheses.}
	\begin{tabular}{lll}
		\hline
		Method & Clean & PGD(20)\\
		\hline
		Normal & $\textbf{73.7}_{(0.1)}$ & $0.0_{(0.0)}$\\
		Hutchinson & $1.0_{(0.0)}$ & $1.0_{(0.0)}$\\
		Hutchinson-0.1 & $69.9_{(0.7)}$ & $5.9_{(0.1)}$\\
		Power Method & $37.4_{(0.7)}$ & $20.2_{(0.7)}$\\
		\textbf{Lanczos (Ours)} & $38.2_{(0.2)}$ & $\textbf{21.1}_{(0.3)}$\\
		\hline
	\end{tabular}
	\end{center}
\end{table}

\subsection{Comparison with Prior Works}
\textbf{Jacobian Regularization.} In Table 1 and Table 2, we compare our Lanczos-based spectral norm minimization with normal training, Hutchinson's estimator, and Power Method. The results show that our technique performs consistently better on all datasets.

Normal training by itself provides the best clean accuracy, however it does not provide any adversarial robustness.
Hutchinson's estimator provides 30.1\% robust accuracy at the cost of a low 60.1\% clean accuracy. It provides a weak trade-off between clean accuracy and robust accuracy compared with our Lanczos-based methodology.
Notably, in the case of CIFAR-100, Hutchinson's estimator is too unstable to provide a meaningful clean or robust accuracy.
On both CIFAR-10 and CIFAR-100, Power Method achieves performance on par with Lanczos algorithm. We believe this is primarily because the Jacobian matrices in these experiments are small. Specifically, it is 10 by 10 for CIFAR-10 and 100 by 100 for CIFAR-100. Significant discrepancy is observed in Hessian regularization, where the Hessian matrices has a constant size of 3072 by 3072.
Our Lanczos-based method consistently achieves a performance gain of 0.1\% and 0.9\% on both datasets.

\textbf{Hessian Regularization.} In Table 3 and Table 4, we compare our Lanczos-based spectral norm minimization with normal training, Hutchinson's estimator, and Power Method. The results show that our technique surpasses other methods by a large margin.

Similar to Jacobian regularization, Hutchinson's estimator provides subpar performance compared with Power Method and Lanczos-based spectral norm minimization. Although Hutchinson-0.1 provides a higher clean accuracy, its robust accuracy is significantly lower than that of spectral norm-based methods.

Although in Jacobian regularization, Power Method provides similar performance compared with the Lanczos algorithm, in the context of Hessian regularization its performance is significantly lower. We believe it is primarily because under Hessian regularization, the matrix is magnitudes larger than that of Jacobian regularization. In this case, Power Method is not converging as fast and accurately as the Lanczos algorithm. 
Our Lanczos-based method consistently achieves a performance gain of 3.7\% and 2.3\% on both datasets.

\begin{table} [!t]
	\caption{Experiment results for Hessian regularization on CIFAR-10. Hutchinson-0.1 means the regularization power is reduced by a factor of 10. The results are averaged over three runs and the standard deviations are reported in parentheses.}
	\begin{center}
	\begin{tabular}{lll}
		\hline
		Method & Clean & PGD(20)\\
		\hline
		Normal & $\textbf{93.3}_{(0.1)}$ & $0.0_{(0.0)}$\\
		Hutchinson & $38.5_{(5.8)}$ & $30.0_{(4.4)}$\\
		Hutchinson-0.1 & $84.8_{(2.0)}$ & $19.5_{(0.1)}$\\
		Power Method & $70.2_{(1.1)}$ & $34.8_{(1.2)}$\\
		\textbf{Lanczos (Ours)} & $73.4_{(0.1)}$ & $\textbf{38.5}_{(0.2)}$\\
		\hline
	\end{tabular}
	\end{center}
\end{table}

\begin{table} [!t]
	\caption{Experiment results for Hessian regularization on CIFAR-100. Hutchinson-0.1 means the regularization power is reduced by a factor of 10. The results are averaged over three runs and the standard deviations are reported in parentheses.}
	\begin{center}
	\begin{tabular}{lll}
		\hline
		Method & Clean & PGD(20)\\
		\hline
		Normal & $\textbf{73.7}_{(0.1)}$ & $0.0_{(0.0)}$\\
		Hutchinson & $13.4_{(0.5)}$ & $8.2_{(0.5)}$\\
		Hutchinson-0.1 & $61.9_{(0.2)}$ & $10.1_{(0.1)}$\\
		Power Method & $43.1_{(1.1)}$ & $17.3_{(0.5)}$\\
		\textbf{Lanczos (Ours)} & $46.2_{(0.5)}$ & $\textbf{19.6}_{(0.4)}$\\
		\hline
	\end{tabular}
	\end{center}
\end{table}

\begin{table} [!t]
	\caption{The running time in seconds per epoch for each method. The task is Hessian regularization on CIFAR-10. Stage 1 is from epoch 1 to 50, Stage 2 is from epoch 51 to 70, Stage 3 is from epoch 71 to 90, and Stage 4 is from epoch 91 to 100.}
	\begin{center}
	\begin{tabular}{lcccc}
		\hline
		Method & Stage 1 & Stage 2 & Stage 3 & Stage 3\\
		\hline
		Hutchinson & 60.2 & 60.2 & 60.2 & 60.2\\
		Power Method & 89.0 & 117.9 & 175.6 & 290.8\\
		Lanczos & 89.1 & 117.9 & 175.6 & 291.1\\
		\hline
	\end{tabular}
	\end{center}
\end{table}

\subsection{Running Time Analysis}
In Table 5 we document the running time for each of the method in our experiments. The running time is recorded on a single NVIDIA A100 GPU. The task is Hessian regularization on the CIFAR-10 dataset.

We use the Hutchinson's method as a baseline because it uses a random vector and do not spend extra time on finding a suitable vector to perform the HVP calculation \citep{fastHVP}. We also note that, as mentioned in Sec.~4.1, both Power Method and the Lanczos algorithm iterates 2, 4, 8, and 16 times for Stage 1, 2, 3, and 4 respectively. 

From Table 5, we draw the following conclusions. First, Power Method and the Lanczos algorithm have identical time costs. Second, for each epoch, the additional time cost introduced by the Lanczos algorithm is $14.4 n$ seconds, where $n$ is the iteration number. Third, depending on the iteration number, using the Lanczos algorithm introduces an overhead ranging from 48\% to 385\%. In total, there is an 120\% overhead. Considering the performance gain provided by the Lanczos algorithm, it is an acceptable cost.

\section{Conclusion}
In this work we generalize the task of regularizing the \mbox{Jacobian} and Hessian matrices of neural networks. Our new paradigm not only permits arbitrary target matrices, but also allows us to explore novel regularizers that enforce symmetry or diagonality for square matrices. Further, we propose Lanczos-based spectral norm minimization, an effective technique for Jacobian and Hessian regularization. We use extensive experiments to validate the effectiveness of our novel regularization terms and the proposed algorithm. Future work includes exploring the possibility of applying the novel regularization terms on Energy-based Models that directly predicts gradient vector fields, thereby ensuring the theoretical integrity.

\bibliography{main}

\appendix
\onecolumn

\section{Gradient Ascent and Power Method at Finding Spectral Norm}
For any matrix $\mathbf{A}$, to find the unit vector that correspond to the spectral norm of $\mathbf{A}^T \mathbf{A}$, one may use Power Method. It is defined by recurrence 
\begin{equation*}
\begin{aligned}
&\mathrm{Initialize~} \mathbf{v}_{1} \mathrm{~as~a~random~unit~vector}\\
&\mathrm{1:}~~\mathbf{v}_{i+1} = \mathbf{A}^T \mathbf{A} \mathbf{v}_{i}\\
&\mathrm{2:}~~\mathbf{v}_{i+1} = \frac{\mathbf{v}_{i+1}}{\Vert \mathbf{v}_{i+1} \Vert_2}\\
&\mathrm{3:}~~\mathrm{repeat~if}~i \leq \mathrm{iteration~number}.
\end{aligned}
\end{equation*}
\citet{hessRegGD} however propose to use gradient ascent to find the $\mathbf{v}$ that maximizes $\Vert \mathbf{A} \mathbf{v}_i \Vert_2$. Given step size $\alpha$, this method is defined by recurrence
\begin{equation*}
\begin{aligned}
&\mathrm{Initialize~} \mathbf{v}_{1} \mathrm{~as~a~random~unit~vector}\\
&\mathrm{1:}~~\mathbf{v}_{i+1} = \mathbf{v}_{i} + \alpha * \nabla_{\mathbf{v}_i}\Vert \mathbf{A} \mathbf{v}_i \Vert_2\\
&\mathrm{2:}~~\mathbf{v}_{i+1} = \frac{\mathbf{v}_{i+1}}{\Vert \mathbf{v}_{i+1} \Vert_2}\\
&\mathrm{3:}~~\mathrm{repeat~if}~i \leq \mathrm{iteration~number}.
\end{aligned}
\end{equation*}
In this section, we show that this method is closely related to Power Method, and they can be practically equivalent. 

We first notice that
\begin{equation}
\nabla_\mathbf{v} \Vert \mathbf{A} \mathbf{v} \Vert_2 = \nabla_\mathbf{v}\sqrt{ \mathbf{v}^T \mathbf{A}^T \mathbf{A} \mathbf{v}} = \frac{1}{\Vert \mathbf{Av} \Vert_2} \mathbf{A}^T \mathbf{Av}.
\end{equation}
Eq.~(1) has two implications. First, to find the $\mathbf{v}$ that maximizes $\Vert \mathbf{A} \Vert_2$, there is no need to to differentiate $\Vert \mathbf{A} \Vert_2$ with respect to $\mathbf{v}$. It suffices to instead perform matrix-vector products. Second, vector $\frac{1}{\Vert \mathbf{Av} \Vert_2} \mathbf{A}^T \mathbf{Av}$ is proportional to $\mathbf{A}^T \mathbf{A} \mathbf{v}$, this strongly relates to the first step of Power Method.

To elaborate more on the second implication, we formulate
$$
\mathbf{v}_{i+1} = \mathbf{v}_{i} + \alpha * \nabla_{\mathbf{v}_i}\Vert \mathbf{A} \mathbf{v}_i \Vert_2
$$
as
$$
\mathbf{v}_{i+1} = \mathbf{v}_{i} + \alpha * \frac{1}{\Vert \mathbf{Av}_{i} \Vert_2} \mathbf{A}^T \mathbf{Av}
$$
Next, we note that $\Vert \mathbf{A}\mathbf{v}_i \Vert_2 \leq \Vert \mathbf{A} \Vert_2$, where $\Vert \mathbf{A} \Vert_2$ is a constant value once $\mathbf{A}$ is fixed. Therefore, suppose that we choose $\alpha$ to be sufficiently large, for example $\alpha = 9 \Vert \mathbf{A} \Vert_2$. Even in the extreme case where $\mathbf{v}_i$ is orthogonal to $\mathbf{A}^T \mathbf{A}\mathbf{v}_i$, after normalization, $\mathbf{v}_{i+1}$ still has a significant cosine similarity of at least $90\%$ with $\frac{\mathbf{A}^T \mathbf{A} \mathbf{v}_i}{\Vert \mathbf{A}^T \mathbf{A} \mathbf{v}_i \Vert_2}$. To conclude, from a theoretical perspective, we believe there is no obvious reason to promote gradient ascent over Power Method.

\vfill
\section{Proof of Theorem 1}

In this section, we give the proof of Theorem 1.
\begin{theorem1}
$\forall{\mathbf{A}} \in \mathbb{R}^{N \times N}$, the following holds
$$\frac{1}{N} \lVert \mathbf{A}-\mathcal{D}(\mathbf{A}\mathbf{1})\rVert_F^2 \leq \sum_i \sum_{j \neq i} \mathbf{A}_{ij}^2 \leq \lVert \mathbf{A}-\mathcal{D}(\mathbf{A}\mathbf{1})\rVert_F^2,$$
where $\mathbf{1} \in \mathbb{R}^{N}$ is an all-one vector, and $\mathcal{D}$ is a function that transforms a vector into a diagonal matrix.
\end{theorem1}

\begin{proof}
Suppose that
$$\mathbf{A}=\begin{bmatrix}\mathbf{a}_1 \hdots \mathbf{a}_N\end{bmatrix}^T,$$ $$\mathbf{B} = \mathbf{A} - \mathcal{D}(\mathbf{A1}) = \begin{bmatrix} \mathbf{b}_1 \hdots \mathbf{b}_N \end{bmatrix}^T, \mathrm{and}$$
$$\mathbf{C} = \mathbf{A} - \mathcal{D}(\mathrm{diag}(\mathbf{A1})) = \begin{bmatrix} \mathbf{c}_1 \hdots \mathbf{c}_N \end{bmatrix}^T.$$
It suffices to show
$$
\frac{1}{N} \Vert \mathbf{B} \Vert_F^2 \leq
\Vert \mathbf{C} \Vert_F^2 \leq
\Vert \mathbf{B} \Vert_F^2.
$$

We observe
$\mathbf{b}_i = \mathbf{a}_i - (\mathbf{a}_i^T\mathbf{1})\mathbf{e}_i$ and $\mathbf{c}_i = \mathbf{a}_i - (\mathbf{a}_i^T\mathbf{e}_i)\mathbf{e}_i$, where $\mathbf{e}_i$ has value 1 at $i^\mathrm{th}$ entry and has value 0 at other entries.
Consequently, 
$\mathbf{b}_i = \mathbf{c}_i + (\mathbf{a}_i^T\mathbf{e}_i)\mathbf{e}_i - (\mathbf{a}_i^T\mathbf{1})\mathbf{e}_i = \mathbf{c}_i - \mathbf{a}_i^T(\mathbf{1} - \mathbf{e}_i)\mathbf{e}_i.$
We note that $\mathbf{1} - \mathbf{e}_i$
has value 0 at $i^\mathrm{th}$ entry and has value 1 at other entries, therefore $\mathbf{a}_i^T(\mathbf{1} - \mathbf{e}_i) = \mathbf{c}_i^T(\mathbf{1} - \mathbf{e}_i)$, and that
$\mathbf{b}_i = \mathbf{c}_i - \mathbf{c}_i^T(\mathbf{1} - \mathbf{e}_i)\mathbf{e}_i$.

Since $\mathbf{c}_i^T \mathbf{e}_i = 0$, $\mathbf{c}_i$ and $\mathbf{c}_i^T(\mathbf{1} - \mathbf{e}_i)\mathbf{e}_i$ are orthogonal, by the Pythagorean theorem and the Cauchy–Schwarz inequality,
$$\Vert \mathbf{b}_i \Vert_2^2 = \Vert \mathbf{c}_i \Vert_2^2 + \Vert \mathbf{c}_i^T(\mathbf{1} - \mathbf{e}_i)\mathbf{e}_i \Vert_2^2 =
\Vert \mathbf{c}_i \Vert_2^2 + \Vert \mathbf{c}_i^T(\mathbf{1} - \mathbf{e}_i) \Vert_2^2 \leq
\Vert \mathbf{c}_i \Vert_2^2 + \Vert \mathbf{c}_i \Vert_2^2 \Vert \mathbf{1} - \mathbf{e}_i \Vert_2^2 =
N \Vert \mathbf{c}_i \Vert_2^2.$$
It is trivial that
$$\Vert \mathbf{b}_i \Vert_2^2 = \Vert \mathbf{c}_i \Vert_2^2 + \Vert \mathbf{c}_i^T(\mathbf{1} - \mathbf{e}_i)\mathbf{e}_i \Vert_2^2 \geq
\Vert \mathbf{c}_i \Vert_2^2.$$

Finally, we have
$$\Vert \mathbf{B} \Vert_F^2 = 
\sum_i \Vert \mathbf{b}_i \Vert_2^2 \geq 
\sum_i \Vert \mathbf{c}_i \Vert_2^2 = 
\Vert \mathbf{C} \Vert_F^2 ~\mathrm{and}$$
$$\frac{1}{N} \Vert \mathbf{B} \Vert_F^2 =
\frac{1}{N} \sum_i \Vert \mathbf{b}_i \Vert_2^2 \leq
\sum_i \Vert \mathbf{c}_i \Vert_2^2 = 
\Vert \mathbf{C} \Vert_F^2.$$
Therefore,
$
\frac{1}{N} \Vert \mathbf{B} \Vert_F^2 \leq
\Vert \mathbf{C} \Vert_F^2 \leq
\Vert \mathbf{B} \Vert_F^2.
$
\end{proof}

\end{document}